\documentclass[10pt,twocolumn,letterpaper]{article}

\usepackage{cvpr}

\usepackage{cuted} 

\usepackage{currfile} 

\usepackage{caption} 

\usepackage{multirow}

\definecolor{cvprblue}{rgb}{0.21,0.49,0.74}
\usepackage[
    pagebackref,
    breaklinks,
    colorlinks,
    allcolors=cvprblue
]{hyperref}

\def\confName{CVPR}
\def\confYear{2027}

\title{
One-Shot Human Video-Guided Adaptation for Vision-Language-Action Models
}

\author{
Qi Chen$^{1,2}$,
Yuwei Wu$^{1,2}$,
Quan Chen$^{1,2,3}$,
Kunpeng Gao$^{1}$, 
Siyuan Ke$^{1}$,\\
Jin Chen$^{3}$,
Che Sun$^{2}$,
Zhi Gao$^{1,2}$,
Yunde Jia$^{2}$%
\thanks{$^{1}$Beijing Key Laboratory of Intelligent Information Technology,
School of Computer Science and Technology,
Beijing Institute of Technology.}%
\thanks{$^{2}$Guangdong Laboratory of Machine Perception and Intelligent Computing,
Shenzhen MSU-BIT University.}%
\thanks{$^{3}$AInnovation Technology Group Co., Ltd.}%
\thanks{Project homepage:
\protect\url{https://chenqi20020226.github.io/cappro-project/}}%
}

\begin{document}

\maketitle


\begin{abstract}
Vision-language-action (VLA) models have recently made substantial progress in general-purpose robot manipulation.
However, rapid adaptation to manipulation tasks beyond the training distribution remains challenging, often requiring target-task robot demonstrations and fine-tuning, which leads to high data and computational costs.
In this paper, we propose CapPro, a framework that combines a phase-aware human video representation module and a video guidance action head with a pretrained VLA policy, enabling the VLA policy to adapt to unseen manipulation tasks given a single human demonstration video at inference. 
The phase-aware human video representation module encodes the demonstration video with temporal progress and extracts execution-context features conditioned on the current robot observation. Phase caption supervision is used during training to encourage the execution-context features to capture task stages. 
The video guidance action head integrates these features
to condition action generation, allowing the human video
to provide task-level execution guidance without explicit
human-to-robot trajectory conversion, strict temporal or
action-level alignment, or target-task robot demonstrations
and fine-tuning.
Experiments on 26 unseen RoboTwin tasks show that CapPro outperforms the Source-trained \(\pi_{0.5}\) baseline and evaluated video-guided baselines, improving target-task average success rate from 10.5\% to 29.5\% while maintaining strong performance on seen  
tasks. Real-world experiments further validate its ability to guide execution on unseen tasks from a single human demonstration video.

\end{abstract}


\section{Introduction}

Vision-language-action (VLA) models~\cite{rt2,openvla,pi0,pi05}
have emerged as a promising paradigm for developing
general-purpose robot manipulation systems by integrating
visual perception, language understanding, and action
generation.
Leveraging large-scale robot datasets~\cite{openx,droid}
and foundation models, recent VLA approaches have
demonstrated strong capabilities across diverse
manipulation tasks~\cite{rt1,octo}.
Despite these advances, pretrained VLA policies do not
readily acquire the task-specific knowledge needed to
perform unseen manipulation tasks.
Adapting them often requires additional robot
demonstrations or policy fine-tuning
~\cite{ricl,dynvla,foca,controlvla,openvla_oft},
increasing data collection and computational costs.

By comparison, humans can often infer how to perform
a new manipulation task from a single video demonstration.
Such videos reveal object interactions and execution
procedures that may not be fully specified by language
instructions or apparent from the robot's current
observation alone.
Prior work has incorporated this information at different
stages of robot learning. During training, human videos
have been used to learn latent plans or video-conditioned
policies, with robot demonstrations or trajectories
providing complementary supervision
~\cite{mimicplay,bharadhwaj2024towards,vid2robot}.
At inference, a single human video can be used to
construct object-centric task plans or functional
representations, or to provide direct execution guidance
~\cite{orion,mimicfunc,host}.
Building on these efforts, we study how a single human
demonstration video can serve as task-specific context
to directly condition a pretrained VLA policy at inference,
without target-task robot demonstrations or fine-tuning.

In doing so, we need to address two challenges.
First, human videos contain substantial visual redundancy,
with task-relevant cues shifting across execution stages.
Coarse task-level representations can lose fine-grained
temporal dynamics, making it challenging to extract and
represent stage-dependent cues for effective robot guidance.
Second, differences in embodiment, observation perspective,
and action space create a gap between human demonstrations
and robot execution, making it difficult to translate
human interactions into executable robot actions.

\begin{figure}[t]
    \centering
    \includegraphics[width=\columnwidth]{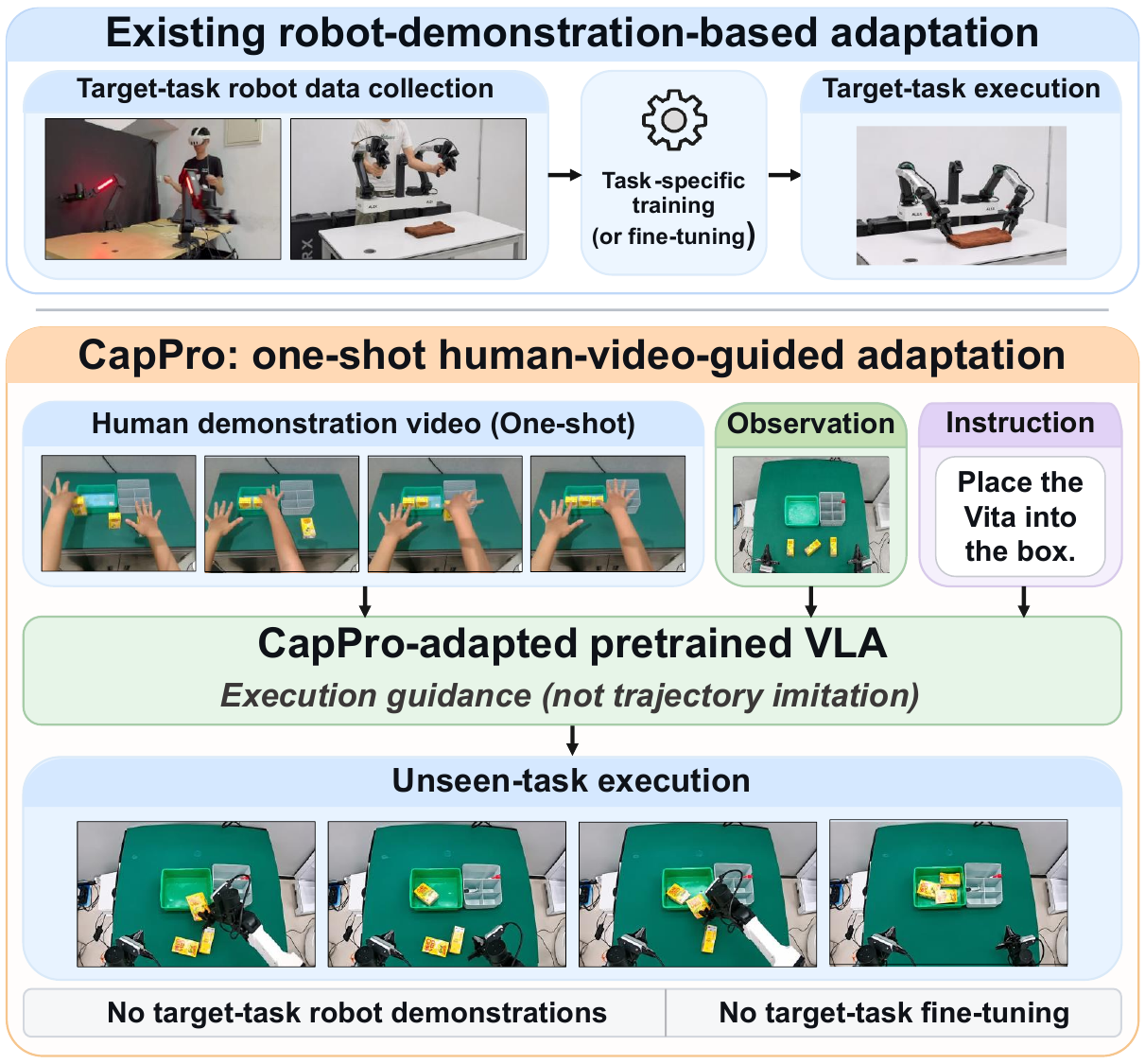}
    \caption{One-shot human-video-guided VLA adaptation.
    Unlike conventional adaptation requiring target-task
    robot demonstrations and fine-tuning (top), CapPro
    uses a single egocentric human video as task-specific
    execution guidance rather than a trajectory to imitate,
    enabling unseen-task execution without additional
    robot demonstrations or fine-tuning.}
    \label{fig:setting}
    \vspace{-0.25in}
\end{figure}

In this paper, we propose CapPro, a framework for one-shot
adaptation of pretrained VLA policies to unseen manipulation
tasks using a single human demonstration video, as illustrated
in Fig.~\ref{fig:setting}.
CapPro consists of two complementary components:
a phase-aware human video representation module and a
video guidance action head.
The former extracts a compact video representation and
an execution-context representation conditioned on the
current robot observation.
During training, phase caption supervision encourages
the execution-context representation to capture task-stage
information relevant to robot execution.
The video guidance action head integrates these representations
with the robot observation and task context to condition
the pretrained VLA policy for action generation without
explicit human-to-robot trajectory conversion.

To systematically evaluate one-shot human-video-guided
adaptation, we establish a benchmark based on
RoboTwin 2.0~\cite{robotwin2}, comprising 39 tasks organized
into 13 source tasks and 26 held-out target tasks across
three levels of task shift.
The benchmark includes human demonstration videos and a
standardized evaluation protocol under clean and randomized
scene configurations, enabling systematic assessment of
adaptation to unseen manipulation tasks.

Our main contributions are summarized as follows:
\begin{itemize}
    \item We propose CapPro, a framework that enables
    one-shot adaptation of pretrained VLA policies to
    unseen manipulation tasks using a single human
    demonstration video, without target-task robot
    demonstrations or fine-tuning.

    \item We establish a RoboTwin-based benchmark and
    standardized evaluation protocol for one-shot
    human-video-guided adaptation.

    \item We conduct extensive simulation and real-world
    experiments to evaluate CapPro. On 26 unseen RoboTwin
    tasks, CapPro improves the target-task average success
    rate from 10.5\% to 29.5\% over the Source-trained
    $\pi_{0.5}$ baseline, while real-world experiments
    further demonstrate its applicability to unseen
    manipulation tasks.
\end{itemize}

\section{Related Work}



\subsection{Vision-Language-Action Models}

Vision-language-action (VLA) models formulate robot
manipulation as a unified mapping from multimodal
observations and language instructions to robot actions.
Large-scale pretraining across diverse tasks and robot
embodiments enables these models to acquire reusable
manipulation capabilities~\cite{bridgedatav2,openx,droid,octo}.
Meanwhile, VLA architectures employ different action modeling
paradigms, ranging from discrete action-token
prediction~\cite{rt1,rt2,openvla} to continuous action-chunk
generation with diffusion- or flow-based
objectives~\cite{rdt1b,pi0,pi05}.
Building on pretrained VLA policies, our work
investigates how execution context extracted from
human demonstrations can provide task-specific
guidance during deployment.

\begin{figure*}[t]
    \centering
    \includegraphics[width=0.96\textwidth]{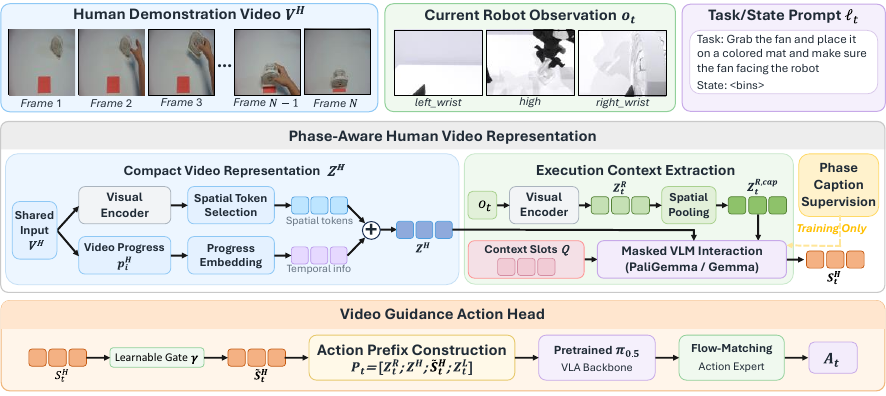}
    \caption{Overview of CapPro. The Phase-Aware Human
    Video Representation module encodes a single human
    demonstration into compact video features $\mathbf{Z}^{H}$
    and extracts execution context $\mathbf{S}_{t}^{H}$
    conditioned on the current robot observation. The Video
    Guidance Action Head integrates both representations
    into the pretrained $\pi_{0.5}$ policy for action
    generation. Phase caption supervision is used only
during training.}
    \label{fig:overview}
\end{figure*}

\subsection{Robot Adaptation}

Adapting pretrained robot policies to unseen tasks
requires task-specific information beyond that acquired
during pretraining.
Existing approaches update policy parameters using
task-specific robot demonstrations through full or
parameter-efficient fine-tuning~\cite{openvla_oft,controlvla,coral,foca},
condition policies on robot demonstrations or retrieved
trajectories~\cite{ricl,recap,ravla}, or reuse structured
skill representations and specialized
modules~\cite{mosvla,dynvla,moira}.
Test-time adaptation further explores policy refinement
through online interaction and optimization at
deployment~\cite{ttvla}.
These approaches generally rely on robot-side experience,
existing skill or trajectory resources, or target-time
optimization.
In contrast, CapPro leverages human demonstration videos
as an alternative source of task-specific information
without target-task robot demonstrations or additional
policy optimization.

\subsection{Learning from Human Videos}

Human videos have been explored as a source of transferable task knowledge for robot manipulation. Existing approaches transfer such knowledge through motion correspondence between human and robot trajectories~\cite{dexmv,egovla,immimic}, intermediate representations of object interactions and task structures~\cite{im2flow2act,track2act,rsrd}, or video-conditioned robot policies~\cite{mimicplay,vid2robot}. Recent studies further investigate one-shot skill acquisition from a single human demonstration~\cite{mimicfunc,orion,host,human2sim2robot}. Rather than explicitly converting human motion into robot actions or learning a separate task-specific policy, CapPro uses a single human video as task-specific context to condition an existing pretrained VLA policy at inference, without requiring explicit trajectory conversion, strict temporal or action-level alignment, or target-task robot demonstrations and fine-tuning.

\section{Method}

\subsection{Problem Formulation}
\label{sec:problem_formulation}

Let $\pi_{\theta}$ denote a pretrained vision-language-action (VLA)
policy. At each execution step $t$, the policy receives the current robot
visual observation $o_t$ and a prompt $\ell_t$ that combines the
language instruction with the discretized current robot state, and
predicts an action chunk
$\mathbf{A}_t=(a_t,a_{t+1},\ldots,a_{t+H-1})$:
\begin{equation}
    \mathbf{A}_t
    \sim
    \pi_{\theta}(\mathbf{A}_t \mid o_t,\ell_t),
\end{equation}
where $H$ denotes the action horizon.

We study one-shot adaptation from source manipulation tasks
to target tasks that are held out during source-task training.
During training, the policy is trained using robot demonstrations
and task-associated human videos from the source tasks.
Although the two modalities share task-level semantics,
they do not require temporal or action-level alignment.
At deployment, the policy is provided with a single human
demonstration video
$V^{H}=(I^{H}_1,I^{H}_2,\ldots,I^{H}_N)$ depicting the target task,
where $I_i^H$ denotes the $i$-th frame of the video and
$N$ is the total number of frames.
The video provides information about task execution, but
is not accompanied by robot actions or trajectories directly
aligned with the robot embodiment.

Conditioning the policy on the human demonstration, the resulting
action generation can be formulated as
\begin{equation}
    \mathbf{A}_t
    \sim
    \pi_{\theta}
    (\mathbf{A}_t \mid o_t,\ell_t,V^{H}).
\end{equation}

\subsection{Framework Overview}

We instantiate CapPro on top of a pretrained
$\pi_{0.5}$ policy. As illustrated in Fig.~2,
CapPro consists of two components: a
Phase-Aware Human Video Representation module
and a Video Guidance Action Head.

The Phase-Aware Human Video Representation module
extracts task information from a single human
demonstration while accounting for the temporal
progression of the demonstrated procedure. It
further relates the demonstrated task to the current
robot observation, allowing the representation to
emphasize information relevant to the current stage
of execution. Together, these representations provide
complementary task-level and state-conditioned guidance.

The Video Guidance Action Head uses the resulting
human video guidance to condition the action
generation process of the pretrained VLA policy.
Rather than directly transferring human motion to
the robot, it integrates task information from the
demonstration with the robot's own observation and
instruction context to guide action prediction.

During training, phase captions supervise the
execution context to capture task-stage information.
At inference, one-shot adaptation is achieved by
conditioning the trained policy on the current robot
observation, prompt $\ell_t$, and a single human
video, without parameter updates or target-task
robot demonstrations. Phase captions and robot
execution progress are used only during training.

\subsection{Phase-Aware Human Video Representation}
\label{sec:human_video_representation}

The Phase-Aware Human Video Representation module extracts task
information from a single human demonstration. A compact representation
captures visual content and temporal progression across the
demonstration, while an execution context representation identifies the
information most relevant to the current robot observation. These two
representations provide complementary guidance for subsequent action
generation.

\paragraph{Compact Human Video Representation.}
Given a human demonstration
$V^H=(I_1^H,\ldots,I_N^H)$, we uniformly sample
$N_s$ frames and encode them using the visual encoder
shared with the pretrained VLA policy. Since the resulting
visual features contain substantial spatial redundancy,
we adopt the differentiable spatial sampling strategy
of GridS~\cite{pmlr-v306-feng26x} to obtain a compact
representation. Specifically, we predict $S$ spatial
sampling coordinates from each frame's visual features
and extract the corresponding tokens through bilinear
interpolation, yielding
$\mathbf{G}_i^H=(g_{i,1}^H,\ldots,g_{i,S}^H)$.

In addition to visual content, the temporal position of each
frame is important for understanding the progression of the
demonstrated task. For the $i$-th sampled frame, let $j_i$
denote the zero-based index of the original frame nearest
to its uniform sampling position, where the first frame is
indexed as $0$. We define its normalized video progress as
\begin{equation}
    p_i^H=\frac{j_i}{N-1}.
\end{equation}
The progress value is encoded using a sinusoidal positional
encoding followed by an MLP, producing a progress embedding
$e_i^H$. This embedding is added to the selected visual
features from the corresponding frame, yielding the compact
human video representation
\begin{equation}
    \mathbf{Z}^{H}
    =
    \left[
    \left(g_{1,s}^{H}+e_1^{H}\right)_{s=1}^{S};
    \ldots;
    \left(g_{N_s,s}^{H}+e_{N_s}^{H}\right)_{s=1}^{S}
    \right].
\end{equation}

\paragraph{Execution Context Representation.}
The compact human video representation contains task information across
the demonstration, while only part of this information may be relevant
to the robot at a particular execution step. We therefore introduce an
execution context extractor that relates the human demonstration to the
current robot observation.

The visual features of the current robot observation are spatially
pooled to obtain a compact robot representation
$\mathbf{Z}_{t}^{R,\mathrm{cap}}$. The extractor introduces $M$
learnable latent slots $\mathbf{Q}=(q_1,\ldots,q_M)$, which are jointly
processed with the human video representation $\mathbf{Z}^{H}$ and the
robot representation $\mathbf{Z}_{t}^{R,\mathrm{cap}}$ through the
pretrained VLM backbone. The contextualized features at these latent
slots form the execution context representation
$\mathbf{S}_{t}^{H}$.

The resulting execution context representation
$\mathbf{S}_{t}^{H}$ is conditioned on the human demonstration
and the current robot observation, providing guidance relevant
to the ongoing execution. During training, the latent slots serve as
the visual information bottleneck for phase caption
prediction, encouraging $\mathbf{S}_{t}^{H}$ to capture
execution stage information, as described in
Sec.~\ref{sec:training_objective}.
At inference time, $\mathbf{S}_{t}^{H}$ is obtained
directly from the human demonstration, the current robot
observation, and the learned latent slots.

\subsection{Video Guidance Action Head}
\label{sec:video_guidance}

The Video Guidance Action Head incorporates the
representations obtained in
Sec.~\ref{sec:human_video_representation}
into the action context of the pretrained VLA policy.
The compact human video representation $\mathbf{Z}^{H}$
provides visual and temporal information from the
demonstration, while the execution context representation
$\mathbf{S}_{t}^{H}$ provides guidance conditioned on
the current robot observation.

To regulate the contribution of the execution context
representation, we introduce a learnable gate:
\begin{equation}
    \widetilde{\mathbf{S}}_{t}^{H}
    =
    \tanh(\gamma)\mathbf{S}_{t}^{H},
\end{equation}
where $\gamma$ is a learnable scalar. The gate allows
the policy to adjust the strength of the execution
context guidance during training.

Let $\mathbf{Z}_{t}^{R}$ denote the robot observation
representation used for action generation in the
pretrained VLA policy, and let $\mathbf{Z}_{t}^{L}$
denote the task and state representation encoded
from the prompt $\ell_t$.
We construct the action prefix as
\begin{equation}
    \mathbf{P}_{t}
    =
    \left[
        \mathbf{Z}_{t}^{R};
        \mathbf{Z}^{H};
        \widetilde{\mathbf{S}}_{t}^{H};
        \mathbf{Z}_{t}^{L}
    \right].
\end{equation}

The augmented action prefix $\mathbf{P}_{t}$ is
provided to the action generation pathway of the
pretrained VLA policy. In this way, CapPro incorporates
human video guidance through the conditioning context
while retaining the original action generation
formulation of the base policy.

\begin{figure}[t]
    \centering
    \includegraphics[width=\columnwidth]{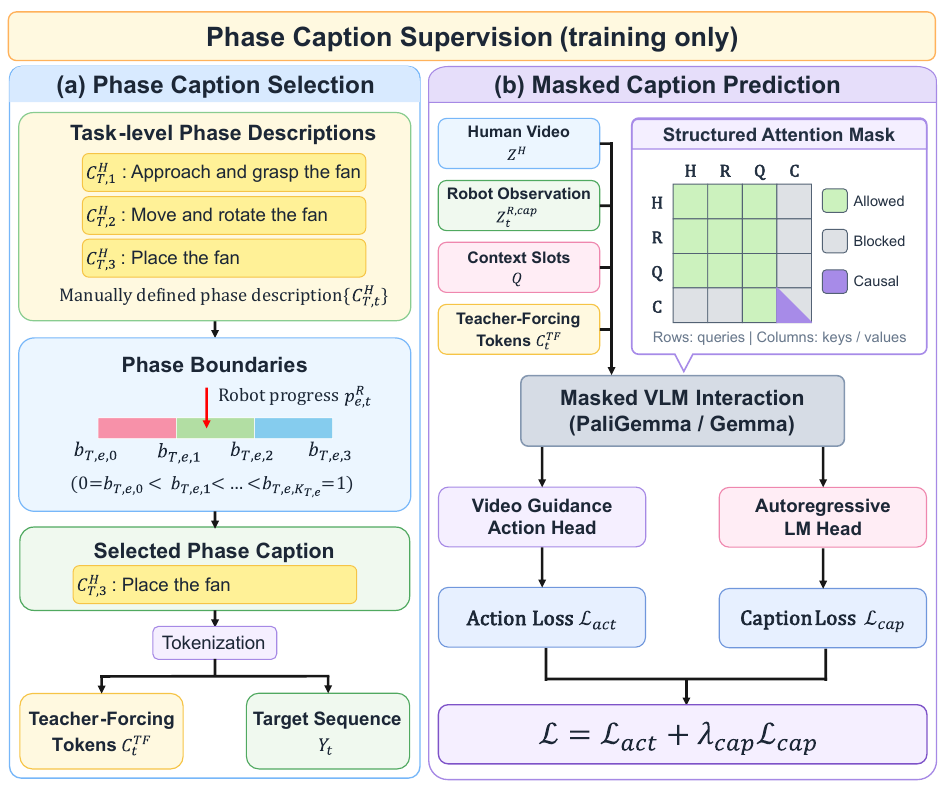}
    \caption{Training-time phase caption supervision. (a) Robot execution progress selects a phase caption from the associated human demonstration using episode-specific phase boundaries. (b) A structured attention mask restricts caption prediction to visual information accessed through the latent context slots, while preventing caption information from influencing these slots. The caption and action generation objectives are jointly optimized.}
    \label{fig:phase_supervision}
\end{figure}

\subsection{Training Objective}
\label{sec:training_objective}

CapPro is trained with two complementary objectives:
the action generation objective of the base VLA policy
and an auxiliary phase caption supervision objective.
The former supervises robot action generation under
human-video guidance, while the latter encourages the
execution context representation to capture information
associated with different stages of task execution.

\paragraph{Action Generation Objective.}
We retain the flow matching objective of the base VLA policy. Given a
ground truth action chunk $\mathbf{A}_{t}^{\mathrm{GT}}$, Gaussian noise
$\boldsymbol{\epsilon}$, and flow time $\tau\in[0,1]$, the noisy action
state is
$\mathbf{A}_{t}^{\tau}
=(1-\tau)\mathbf{A}_{t}^{\mathrm{GT}}
+\tau\boldsymbol{\epsilon}$.
Conditioned on the video augmented action prefix
$\mathbf{P}_{t}$ defined in
Sec.~\ref{sec:video_guidance}, the action loss is
\begin{equation}
    \mathcal{L}_{\mathrm{act}}
    =
    \mathbb{E}_{\tau,\boldsymbol{\epsilon}}
    \left[
        \left\|
        v_{\theta}
        \left(
            \mathbf{A}_{t}^{\tau},
            \tau
            \mid
            \mathbf{P}_{t}
        \right)
        -
        \left(
            \boldsymbol{\epsilon}
            -
            \mathbf{A}_{t}^{\mathrm{GT}}
        \right)
        \right\|_{2}^{2}
    \right].
\end{equation}
Here, $v_{\theta}$ denotes the velocity field predicted
by the action expert.

\paragraph{Phase Caption Supervision.}
For each source task $\mathcal{T}$, we manually divide the
manipulation procedure into $K_{\mathcal{T}}$ semantically
meaningful phases. 
Each human demonstration of task
$\mathcal{T}$ is associated with a set of phase descriptions
$\mathcal{C}_{\mathcal{T}}^{H}
=\{c_{\mathcal{T},1}^{H},\ldots,
c_{\mathcal{T},K_{\mathcal{T}}}^{H}\}$,
which describe the corresponding stages of execution
in that video. For notational simplicity, the human
demonstration index is omitted from the phase caption
notation. For example, the \textit{place fan} task
consists of three phases:
\textit{``Approach and grasp the fan''},
\textit{``Move and rotate the fan''}, and
\textit{``Place the fan''}.
While the phase structure is shared within a task,
the descriptions reflect the specific human demonstration,
such as the hand used for manipulation.

As illustrated in Fig.~\ref{fig:phase_supervision},
the phase caption used for training is selected according to
the current progress of the robot episode. Since different
episodes may execute the same task at different rates, we
determine episode-specific phase boundaries offline using
task-specific kinematic rules applied to robot joint
trajectories. For episode $e$ of task $\mathcal{T}$,
the normalized boundaries are denoted as
\begin{equation}
    0=b_0^{\mathcal{T},e}
    <b_1^{\mathcal{T},e}
    <\cdots
    <b_{K_{\mathcal{T}}}^{\mathcal{T},e}=1.
\end{equation}

For a training sample at step $t$, its normalized robot
progress $p_{e,t}^{R}$ determines the current phase $k$
such that
$p_{e,t}^{R}\in
[b_{k-1}^{\mathcal{T},e},b_k^{\mathcal{T},e})$,
with the final interval including the endpoint at $1$.
The corresponding caption $c_{\mathcal{T},k}^{H}$
from the human demonstration associated with the
current robot training sample is then used as
auxiliary supervision. In this way, phase supervision
is established through the shared task structure without
requiring temporal alignment between human and robot
demonstrations. Robot progress is used only to select
the training caption and is not provided as an input
to the policy.

During training, the selected phase description is tokenized
to construct an input sequence $\mathbf{C}_{t}^{\mathrm{TF}}$
for teacher forcing. The human video representation
$\mathbf{Z}^{H}$, the compact robot representation
$\mathbf{Z}_{t}^{R,\mathrm{cap}}$, the latent slots
$\mathbf{Q}$, and $\mathbf{C}_{t}^{\mathrm{TF}}$ are jointly
processed by the VLM backbone.

We employ a structured attention mask to establish an
information bottleneck for phase caption prediction.
The visual tokens and latent slots can attend to each
other, but neither can attend to the caption tokens.
Conversely, the caption tokens can attend to the latent
slots and preceding caption tokens, but cannot directly
access the visual tokens. This design ensures that
phase caption prediction obtains visual information
through the execution context representation
$\mathbf{S}_{t}^{H}$, while preventing the ground-truth
caption from influencing this representation.

Let $(y_1,\ldots,y_L,y_{L+1})$ denote the
target token sequence of the selected phase description,
where $y_{L+1}=y_{\mathrm{EOS}}$ is the end-of-sequence token.
For a single caption, the autoregressive language
modeling loss is defined as
\begin{equation}
    \mathcal{L}_{\mathrm{cap}}
    =
    -\frac{1}{L+1}
    \sum_{m=1}^{L+1}
    \log
    p_{\theta}
    \left(
        y_m
        \mid
        y_{<m},
        \mathbf{Z}_{t}^{R,\mathrm{cap}},
        \mathbf{Z}^{H},
        \mathbf{Q}
    \right).
\end{equation}

The overall training objective is
\begin{equation}
    \mathcal{L}
    =
    \mathcal{L}_{\mathrm{act}}
    +
    \lambda_{\mathrm{cap}}
    \mathcal{L}_{\mathrm{cap}},
\end{equation}
where $\lambda_{\mathrm{cap}}$ controls the contribution
of the auxiliary phase supervision.

\section{Experiments}

\subsection{Benchmark}

\noindent\textbf{Task Splits.}
We instantiate our source-to-target adaptation setting
on RoboTwin 2.0~\cite{robotwin2}. As summarized in
Table~\ref{tab:task_split1}, we organize 39 manipulation
tasks into four splits: 13 Source, 6 Target-A1,
10 Target-A2, and 10 Target-B tasks.
The Source tasks are used for training and seen-task
evaluation, while the 26 Target tasks are held out
during training to evaluate adaptation to unseen tasks.
Target-A1 involves relatively small changes from the
Source tasks, Target-A2 introduces larger changes in
objects, interactions, or affordances, and Target-B
involves compositional manipulation tasks.

\begin{table}[t]
    \centering
    \caption{RoboTwin task splits and human video
    distribution. C/R denotes clean/randomized
    videos per task.}
    \label{tab:task_split1}
    \footnotesize
    \setlength{\tabcolsep}{2pt}
    \renewcommand{\arraystretch}{1.12}
    \begin{tabular}{@{}l c c c l@{}}
        \hline
        \textbf{Split} &
        \textbf{\# Tasks} &
        \textbf{Train C/R} &
        \textbf{Eval C/R} &
        \textbf{Task Shift} \\
        \hline
        Source    & 13 & 5/5 & 10/10 & Training tasks \\
        Target-A1 & 6  & --  & 10/10 & Near-task variations \\
        Target-A2 & 10 & --  & 10/10 & Object/interaction/affordance \\
        Target-B  & 10 & --  & 10/10 & Compositional tasks \\
        \hline
    \end{tabular}
\end{table}

\noindent\textbf{Scene Configurations.}
RoboTwin provides two scene configurations:
clean and randomized. The clean configuration uses
a fixed background, lighting, and tabletop setup
without distractors, while the randomized configuration
varies background textures, distractor objects,
table height, and lighting.
Object initializations are randomized in both
configurations.

\noindent\textbf{Human Demonstration Videos.}
We record egocentric demonstrations with physical
objects resembling RoboTwin assets from the simulated
head-camera viewpoint. Paired clean/randomized videos
share core object instances and initial poses but differ
in environmental conditions. The demonstrations match
simulation task objectives without temporal or
action-level alignment. Table~\ref{tab:task_split1}
summarizes the 130 Source training and 780 evaluation
videos across 39 tasks, with no overlap. Each rollout
uses one human demonstration video.

\subsection{Experimental Setup}
\label{sec:experimental_setup}

\noindent\textbf{Training Configuration.}
CapPro is initialized from a pretrained $\pi_{0.5}$
policy and trained on 7,150 robot trajectories from
the 13 Source tasks (50 clean and 500 randomized
per task). We apply LoRA to the VLM backbone and
action expert, training for 100K steps with AdamW
and a cosine learning rate schedule. Key
hyperparameters are listed in Table~\ref{tab:hyperparameters}.

\noindent\textbf{Baselines.}
We compare CapPro with six baselines. Four use the
pretrained $\pi_{0.5}$ backbone: a Source-trained
$\pi_{0.5}$ without video guidance, \textit{Direct
Video Prefix}, which concatenates human video tokens,
\textit{Vid2Robot-style}~\cite{vid2robot}, which uses
paired human--robot streams with cross-attention and
temporal alignment, and \textit{OSVI-WM-style}~\cite{osviwm},
which recursively predicts future robot latents.
We also include \textit{HOST}~\cite{host}, which
aligns human--robot progress and predicts future
observations, and \textit{SeeTraceAct}~\cite{seetraceact},
which uses a video-derived latent plan supervised
by future end-effector traces and in-frame validity.
All methods share the training budget and applicable
robot and human video data. A fully supervised
$\pi_{0.5}$ trained on all 39 tasks serves as a reference.

\begin{table}[t]
    \centering
    \caption{Key hyperparameters. LoRA values are
    listed as VLM / action expert.}
    \label{tab:hyperparameters}
    \footnotesize
    \setlength{\tabcolsep}{3pt}
    \renewcommand{\arraystretch}{1.0}
    \begin{tabular*}{\columnwidth}
    {@{\extracolsep{\fill}}lclc@{}}
        \hline
        \textbf{Parameter} & \textbf{Value} &
        \textbf{Parameter} & \textbf{Value} \\
        \hline
        Video frames $N_s$ & 8 &
        Action horizon $H$ & 50 \\
        Tokens/frame $S$ & 32 &
        Batch size & 128 \\
        Context slots $M$ & 4 &
        Peak LR & $2.5\times10^{-5}$ \\
        Caption weight $\lambda_{\mathrm{cap}}$ & 0.1 &
        LoRA rank & 16/32 \\
        Gate init. & 0.1 &
        LoRA alpha & 16/32 \\
        \hline
    \end{tabular*}
\end{table}

\begin{table*}[t]
    \centering
    \caption{Success rates (\%) on RoboTwin 2.0 (mean $\pm$ SD). The last row uses robot demonstrations from all 39 tasks for training.}
    \label{tab:main_results}
    \footnotesize
    \setlength{\tabcolsep}{5pt}
    \renewcommand{\arraystretch}{1.12}

    \begin{tabular*}{\textwidth}{@{\extracolsep{\fill}}lcccccc@{}}
        \hline
        \textbf{Method} &
        \textbf{Source} &
        \textbf{Target-A1} &
        \textbf{Target-A2} &
        \textbf{Target-B} &
        \textbf{Target Avg.} &
        \textbf{Overall} \\
        \hline

        $\pi_{0.5}$ (Source-trained) &
        $64.2 \pm 0.6$ &
        $34.6 \pm 1.3$ &
        $2.4 \pm 0.3$ &
        $4.0 \pm 0.3$ &
        $10.5 \pm 0.4$ &
        $28.4 \pm 0.2$ \\

        Direct Video Prefix &
        $43.7 \pm 3.1$ &
        $25.3 \pm 2.3$ &
        $1.5 \pm 1.4$ &
        $1.4 \pm 1.2$ &
        $6.9 \pm 1.0$ &
        $19.2 \pm 1.1$ \\

        Vid2Robot-style &
        $65.7 \pm 7.2$ &
        $\mathbf{45.7 \pm 8.4}$ &
        $10.7 \pm 2.2$ &
        $4.1 \pm 1.9$ &
        $16.2 \pm 2.0$ &
        $32.7 \pm 2.6$ \\

        OSVI-WM-style &
        $63.3 \pm 3.1$ &
        $43.9 \pm 1.7$ &
        $9.3 \pm 1.9$ &
        $5.5 \pm 1.4$ &
        $15.8 \pm 0.8$ &
        $31.7 \pm 1.4$ \\

        HOST &
        $63.9 \pm 5.2$ &
        $42.6 \pm 3.5$ &
        $16.8 \pm 3.6$ &
        $11.5 \pm 2.2$ &
        $20.7 \pm 1.7$ &
        $35.1 \pm 1.9$ \\

        SeeTraceAct &
        $64.8 \pm 5.1$ &
        $43.2 \pm 3.6$ &
        $17.1 \pm 2.2$ &
        $9.4 \pm 1.9$ &
        $20.2 \pm 1.5$ &
        $35.0 \pm 2.3$ \\

        \textbf{CapPro} &
        $\mathbf{68.6 \pm 4.1}$ &
        $45.0 \pm 4.5$ &
        $\mathbf{29.0 \pm 2.3}$ &
        $\mathbf{20.7 \pm 2.9}$ &
        $\mathbf{29.5 \pm 2.0}$ &
        $\mathbf{42.5 \pm 1.8}$ \\

        \hline

        $\pi_{0.5}$ (Full Supervision) &
        $63.7 \pm 0.9$ &
        $52.9 \pm 1.7$ &
        $49.6 \pm 0.2$ &
        $45.4 \pm 0.2$ &
        $48.7 \pm 0.3$ &
        $53.7 \pm 0.4$ \\

        \hline
    \end{tabular*}
\end{table*}

\noindent\textbf{Evaluation Protocol.}
All methods are evaluated in ten batches of 100
rollouts per task and scene configuration, using
the same rollout seeds across batches. For video-guided methods, each batch uses a fixed
pair of held-out videos, one per configuration,
with one video per rollout. The
Source-trained and fully supervised $\pi_{0.5}$
policies are evaluated without human video input.
Evaluation uses a separate random seed range from
training. Within each batch, clean and randomized
success rates are averaged per task before computing
split-level and overall results. We report the mean
and standard deviation (SD) of these batch-level
results across ten batches. No additional test-time
fine-tuning or policy updates are performed.

\subsection{Simulation Results}

Table~\ref{tab:main_results} summarizes the results
across RoboTwin task splits. Using the same Source
robot demonstrations, CapPro substantially outperforms
the Source-trained $\pi_{0.5}$ on unseen tasks,
demonstrating the benefit of human-video-guided
adaptation. It also achieves the highest target average
among video-guided baselines. CapPro performs
comparably to Vid2Robot-style on Target-A1, while
achieving larger gains on Target-A2 and Target-B,
indicating stronger adaptation under larger task shifts.
Notably, Direct Video Prefix performs worse than the
Source-trained $\pi_{0.5}$, suggesting that simply
concatenating human video tokens is insufficient
for effective adaptation.

For reference, we additionally evaluate a fully
supervised $\pi_{0.5}$ policy trained with robot
demonstrations from all 39 tasks, including the
held-out target tasks. As shown in
Table~\ref{tab:main_results}, CapPro achieves a
target average of 29.5\%, compared with 48.7\%
for the fully supervised reference. The gap is
relatively small on Target-A1 but becomes larger
on Target-A2 and Target-B, highlighting the remaining
challenges in adapting to substantial task shifts
without target-task robot demonstrations.

\noindent\textbf{Failure Analysis.}
We analyze 2,600 failed CapPro rollouts from the
26 unseen target tasks, sampling 50 failures per
task under each scene configuration (clean and randomized).
Failures are categorized into five types based on
their primary execution errors, as shown in
Table~\ref{tab:failure_analysis}. The most frequent
failure type is final-state and goal completion
failure (E5, 33.8\%), where the preceding execution
generally follows the task requirements but the
final state does not satisfy the success criteria.
Grounding (E1) and sequencing (E4) errors account
for 22.1\% and 18.7\%, respectively.
These results highlight remaining challenges in
target grounding, action organization, and satisfying
task-specific final-state constraints, beyond
low-level grasping and motion execution.

\begin{table}[t]
    \centering
    \caption{Failure types across 2,600 sampled failed
    rollouts from 26 unseen tasks, reported as count
    (percentage).}
    \label{tab:failure_analysis}
    \footnotesize
    \setlength{\tabcolsep}{3pt}
    \renewcommand{\arraystretch}{1.0}

    \begin{tabular*}{\columnwidth}
    {@{\extracolsep{\fill}}lccc@{}}
        \hline
        \textbf{Failure Type} &
        \textbf{Clean} &
        \textbf{Randomized} &
        \textbf{Overall} \\
        \hline

        E1: Grounding &
        274 (21.1) & 300 (23.1) & 574 (22.1) \\

        E2: Grasp &
        174 (13.4) & 170 (13.1) & 344 (13.2) \\

        E3: Motion &
        184 (14.2) & 131 (10.1) & 315 (12.1) \\

        E4: Sequencing &
        260 (20.0) & 227 (17.5) & 487 (18.7) \\

        E5: Final-state &
        408 (31.4) & 472 (36.3) & 880 (33.8) \\

        \hline
        Total &
        1300 (100) & 1300 (100) & 2600 (100) \\
        \hline
    \end{tabular*}
\end{table}

\subsection{Ablation Studies}

We investigate three aspects of CapPro: human video
utilization, temporal information, and phase-aware
supervision and execution guidance. Mismatched Video,
First \& Last Frames, and Shuffled Frames modify only
the inference-time video input, while all other
variants are retrained with the corresponding changes.
All variants use the same Source robot training data
and follow the evaluation protocol in
Sec.~\ref{sec:experimental_setup}.
Results are summarized in Table~\ref{tab:ablation}.

\begin{table*}[t]
    \centering
    \caption{Ablation results on RoboTwin (mean $\pm$ SD, \%). Target Avg. is averaged over 26 unseen tasks.}
    \label{tab:ablation}
    \footnotesize
    \setlength{\tabcolsep}{3pt}
    \renewcommand{\arraystretch}{1.0}

    \begin{tabular*}{\textwidth}
    {@{\extracolsep{\fill}}lcccccc@{}}
        \hline
        \textbf{Configuration} &
        \textbf{Source} &
        \textbf{Target-A1} &
        \textbf{Target-A2} &
        \textbf{Target-B} &
        \textbf{Target Avg.} &
        \textbf{Overall} \\
        \hline

        \textbf{Full CapPro} &
        $68.6 \pm 4.1$ &
        $\mathbf{45.0 \pm 4.5}$ &
        $\mathbf{29.0 \pm 2.3}$ &
        $\mathbf{20.7 \pm 2.9}$ &
        $\mathbf{29.5 \pm 2.0}$ &
        $\mathbf{42.5 \pm 1.8}$ \\
        \hline

        \multicolumn{7}{l}{
        \textit{Human Video Utilization}} \\

        Mismatched Video &
        $51.1 \pm 3.1$ &
        $24.9 \pm 2.7$ &
        $3.1 \pm 1.4$ &
        $2.7 \pm 1.3$ &
        $8.0 \pm 1.2$ &
        $22.4 \pm 1.5$ \\

        VLM-Generated Text Guidance &
        $55.8 \pm 5.6$ &
        $34.1 \pm 5.2$ &
        $4.1 \pm 1.3$ &
        $4.0 \pm 1.6$ &
        $11.0 \pm 1.5$ &
        $25.9 \pm 2.5$ \\

        \hline

        \multicolumn{7}{l}{
        \textit{Temporal Information}} \\

        First \& Last Frames &
        $61.4 \pm 2.1$ &
        $37.7 \pm 2.3$ &
        $10.1 \pm 1.2$ &
        $10.0 \pm 1.6$ &
        $16.4 \pm 1.1$ &
        $31.4 \pm 1.2$ \\

        Shuffled Frames &
        $58.9 \pm 3.5$ &
        $36.9 \pm 3.2$ &
        $11.9 \pm 2.1$ &
        $8.7 \pm 1.3$ &
        $16.4 \pm 1.3$ &
        $30.6 \pm 1.5$ \\

        w/o Human Video Progress &
        $62.4 \pm 3.4$ &
        $36.3 \pm 4.2$ &
        $4.7 \pm 1.7$ &
        $2.5 \pm 1.1$ &
        $11.2 \pm 1.3$ &
        $28.2 \pm 1.1$ \\

        \hline

        \multicolumn{7}{l}{
        \textit{Phase-Aware Supervision and Execution Guidance}} \\

        w/o Phase-wise Caption Selection &
        $63.7 \pm 3.6$ &
        $38.8 \pm 2.3$ &
        $6.1 \pm 1.3$ &
        $6.9 \pm 1.5$ &
        $14.0 \pm 1.0$ &
        $30.5 \pm 1.1$ \\

        w/o Caption Loss &
        $\mathbf{69.5 \pm 2.9}$ &
        $40.9 \pm 2.5$ &
        $8.6 \pm 1.8$ &
        $9.8 \pm 1.6$ &
        $16.5 \pm 0.9$ &
        $34.2 \pm 1.0$ \\

        w/o Execution Context Guidance &
        $63.7 \pm 4.2$ &
        $39.9 \pm 4.7$ &
        $7.1 \pm 2.6$ &
        $8.1 \pm 1.9$ &
        $15.1 \pm 1.3$ &
        $31.3 \pm 1.9$ \\

        \hline
    \end{tabular*}
\end{table*}

\noindent\textbf{Human Video Utilization.}
The substantial drop with mismatched videos (8.0\%)
shows that the effectiveness of video guidance relies
on task-level correspondence between human demonstrations
and robot execution. Notably, VLM-generated textual
descriptions achieve only 11.0\%, even though the original
phase caption supervision is retained. This suggests
that describing the demonstrated procedure in language
does not fully preserve the visual cues needed for
robot execution.

\noindent\textbf{Temporal Information.}
The comparable performance of First \& Last Frames
and Shuffled Frames (both 16.4\%) reflects two
distinct limitations: the loss of intermediate
manipulation content and the mismatch between
shuffled frames and their temporal positions.
The separately retrained model without explicit
video progress scores 11.2\%, suggesting that
temporal position encoding provides complementary
information beyond visual content and frame ordering.

\noindent\textbf{Phase-Aware Supervision and Execution
Guidance.}
A single caption describing the entire demonstration
achieves only 14.0\%, compared with 29.5\% under
phase-wise caption selection, suggesting that global
task descriptions are insufficient to capture
execution-stage-specific information. The results
without caption loss (16.5\%) and execution context
guidance (15.1\%) further distinguish the roles of
the two components: caption supervision shapes the
execution context, while its integration into action
generation makes the learned context available for
robot control. Notably, the former retains Source
performance but degrades target-task adaptation.

\begin{figure*}[t]
    \centering
    \includegraphics[width=\textwidth]{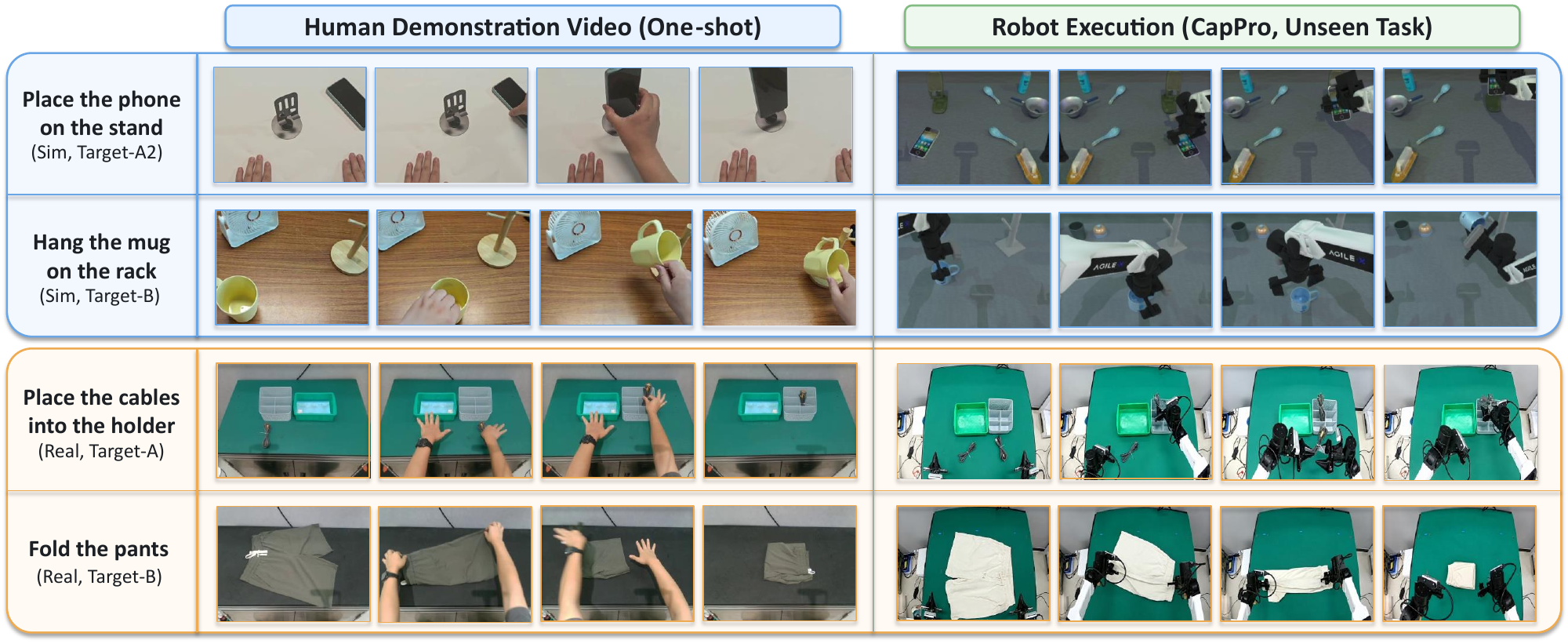}
    \caption{Qualitative results on unseen tasks in
    simulation and the real world. Each row shows a
    single human demonstration (left) and CapPro
    execution (right), without temporal alignment.}
    \vspace{-0.1in}
    \label{fig:qualitative}
\end{figure*}

\begin{table}[t]
    \centering
    \caption{Real-world success rates (\%) over 20 trials
    per task.}
    \label{tab:real_results}
    \footnotesize
    \setlength{\tabcolsep}{4pt}
    \renewcommand{\arraystretch}{1.0}

    \begin{tabular}{@{}l l c c@{}}
        \hline
        \multirow{2}{*}{\textbf{Split}} &
        \multirow{2}{*}{\textbf{Task}} &
        \multirow{2}{*}{\shortstack[c]{\textbf{Source-trained}\\$\pi_{0.5}$}} &
        \multirow{2}{*}{\textbf{CapPro}} \\
        & & & \\
        \hline
        Source &
        Coke $\rightarrow$ Box &
        80 (16/20) & \textbf{90 (18/20)} \\

        Source &
        Marker $\rightarrow$ Holder &
        \textbf{100 (20/20)} &
        \textbf{100 (20/20)} \\

        Source &
        Fold Short-Sleeve Garment &
        85 (17/20) & \textbf{90 (18/20)} \\

        \multicolumn{2}{l}{Source Avg.} &
        88.3 & \textbf{93.3} \\

        \hline

        Target-A &
        Vita $\rightarrow$ Box &
        70 (14/20) & \textbf{100 (20/20)} \\

        Target-A &
        Sprite $\rightarrow$ Box &
        80 (16/20) & \textbf{95 (19/20)} \\

        Target-A &
        Cable Coil $\rightarrow$ Holder &
        \textbf{85 (17/20)} &
        \textbf{85 (17/20)} \\

        \multicolumn{2}{l}{Target-A Avg.} &
        78.3 & \textbf{93.3} \\

        \hline

        Target-B &
        Fold Pants &
        0 (0/20) & \textbf{65 (13/20)} \\

        Target-B &
        Fold Long-Sleeve Sweatshirt &
        0 (0/20) & \textbf{20 (4/20)} \\

        Target-B &
        Fold Towel &
        0 (0/20) & \textbf{30 (6/20)} \\

        Target-B &
        Workpiece Sorting &
        0 (0/20) & \textbf{25 (5/20)} \\

        \multicolumn{2}{l}{Target-B Avg.} &
        0.0 & \textbf{35.0} \\

        \hline

        \multicolumn{2}{l}{Target Avg. (7 tasks)} &
        33.6 & \textbf{60.0} \\

        \multicolumn{2}{l}{Overall Avg. (10 tasks)} &
        50.0 & \textbf{70.0} \\

        \hline
    \end{tabular}
    \vspace{-0.1in}
\end{table}

\subsection{Real-World Experiments}

\noindent\textbf{Experimental Setup.}
We evaluate CapPro on an ARX AC one dual-arm
robotic platform equipped with two X5 robotic arms,
their native grippers, and three RealSense 435
cameras providing overhead, left-wrist, and
right-wrist observations. The policy retains the
same architecture, multi-view observation structure,
and action interface as in simulation.

We consider ten real-world tasks, comprising three
Source, three Target-A, and four Target-B tasks
(Table~6), covering object substitution, garment
folding, and more complex manipulation procedures.
Both CapPro and the Source-trained $\pi_{0.5}$ baseline
are trained on real-world robot demonstrations from
the three Source tasks and evaluated over 20 trials
per task. CapPro uses four human demonstration videos
per task, each fixed for five rollouts, while the
baseline operates without video input. Neither policy
uses target-task robot demonstrations or additional
fine-tuning.

Objects are randomly initialized within the workspace,
and garments are laid flat with randomized orientations.
The long-sleeve sweatshirt follows the short-sleeve
folding criterion, while towel folding requires three
folds along its longer dimension and one along its
shorter dimension. Success requires completion of
the full task objective.

\noindent\textbf{Results.}
As shown in Table~\ref{tab:real_results}, CapPro
improves the average success rate on
unseen tasks from 33.6\% to 60.0\%, while outperforming
the Source-trained $\pi_{0.5}$ baseline on Source and Target-A.
Notably, CapPro achieves 35.0\% on Target-B, where the
baseline fails all four tasks.
Figure~\ref{fig:qualitative} illustrates representative
executions on unseen tasks in simulation and the real world.

\section{Conclusion}

We presented CapPro, a framework for one-shot adaptation
of pretrained VLA policies to unseen manipulation tasks
using a single egocentric human demonstration video.
By extracting phase-aware visual representations and
execution context from human demonstrations, CapPro
provides task-specific guidance for robot action
generation without requiring target-task robot
demonstrations or additional fine-tuning. Experiments
in simulation and the real world demonstrate its
effectiveness in adapting to unseen tasks, while
ablation studies reveal the importance of preserving
task-relevant visual and temporal information and
connecting phase caption supervision to action generation.
Despite these advances, a gap remains compared with
full robot supervision, particularly under substantial
task shifts. Future work will explore more effective
representations and execution guidance for complex
manipulation tasks.

\section*{LLM Usage Statement}

OpenAI ChatGPT was used to assist with manuscript
writing and revision, provide feedback on figure
design and presentation, and verify numerical
calculations. All figures were created by the authors
using Adobe Illustrator, with original images obtained
from author-recorded demonstrations or photographs.
No AI-generated visual content was used. All
experimental data were collected and statistically
reported by the authors. All AI-assisted suggestions
and calculation checks were reviewed by the authors,
who take full responsibility for the final manuscript
and its scientific claims.


{
    \small
    \bibliographystyle{ieeenat_fullname}
    \bibliography{main}
}

\end{document}